# Reverse-engineering Bar Charts Using Neural Networks

Fangfang Zhou[1], Yong Zhao[1], Wenjiang Chen[1], Yijing Tan[1], Yaqi Xu[1], Yi Chen[2], Chao Liu[3], Ying Zhao[1,4*]

[1]School of Computer Science and Engineer, Central South University, Changsha, China;
[2]Beijing Key Laboratory of Big Data Technology for Food Safety, Beijing Technology and Business University, Beijing, China.
[3]Institute of Systems Engineering, Academy of Military Sciences, Beijing, China
[4]Rail Data Research and Application Key Laboratory of Hunan Province, Changsha, China;
*The corresponding author: Ying Zhao, zhaoying@csu.edu.cn

**Abstract**—Reverse-engineering bar charts extracts textual and numeric information from the visual representations of bar charts to support application scenarios that require the underlying information. In this paper, we propose a neural network-based method for reverse-engineering bar charts. We adopt a neural network-based object detection model to simultaneously localize and classify textual information. This approach improves the efficiency of textual information extraction. We design an encoder–decoder framework that integrates convolutional and recurrent neural networks to extract numeric information. We further introduce an attention mechanism into the framework to achieve high accuracy and robustness. Synthetic and real-world datasets are used to evaluate the effectiveness of the method. To the best of our knowledge, this work takes the lead in constructing a complete neural network-based method of reverse-engineering bar charts.

## 1 Introduction

Bar charts, which are popular chart types on the Internet[1], are commonly used to visually present quantitative information. In most cases, only visual representations of the charts but not the underlying data are available. Extracting the underlying raw data is a common requirement in many application scenarios[2, 3]. For example, journalists who are compiling news find some interesting statistics expressed in old-style bar charts and want to use them in their article. Without extracting tools, they must manually extract the raw data to perform chart redesign or further analysis[4]. For another example, a team of software engineers intends to build a chart search engine[5, 6]. They typically require automated tools to extract the raw data, such as the chart title and axis title, to build accurate indexes.

Reverse-engineering bar charts are used to extract chart information. Information in bar charts has two main types: textual and numeric information. Thus, extracting chart information indicates the extraction of textual and numeric information. Textual information annotates the visualization to provide much detail and includes several categories of texts: title, legend, axis title, and axis label (axis includes x-axis and y-axis). Different text types vary in their character strings and placements. Numeric information presents statistical result and is the main component in charts. The bars within bar charts are used to encode numeric information, usually through the height of the bars.

Charts are designed to leverage our vision system; thus, we can easily extract information from them. However, this case is inapplicable to machines. Several prior studies[7, 8] exist to investigate the extraction of textual information from bar charts. They largely follow a bottom-up approach that involves pixel-level classification and merge, word-level detection and merge, and text classification. Such approach is usually inefficient because it involves many low-level operations and slightly excessive procedures. Previous works [2-3,5-8] on numeric information extraction mostly use traditional image processing techniques and task-specific rules. They initially find rectangular shapes, then filter the shapes according to rules, and finally map the shapes back to numeric data. These methods generally fail to achieve high accuracy because of two reasons. First, achieving high accuracy with traditional image processing techniques is difficult for their limited capabilities. Second, fixed rules are vulnerable to variation in graphical elements that encode numeric information. This vulnerability further leads to decreased accuracy.

Neural networks, which are considered a branch of artificial intelligence, feature self-learning ability. Neural networks are generally more efficient and have better performance in many image-related problems, such as image classification[9], object detection[10] and image captioning[11], than traditional image processing techniques. Pix2code[12] is a novel method that uses neural networks to reverse-engineer graphical user interfaces (GUI), from GUI to computer code, and achieves over 77% accuracy. Pix2code's successful attempt demonstrates the possibility of using neural networks for reverse-engineering tasks. There are also a few recent studies that introduce deep learning into chart information extraction. Scatteract[13] uses a neural network-based object detection model to locate the points in a scatterplot and then extracts the numerical information based on the positioning, but this method is not capable for other types of charts. Liu[14] et al. used deep neural networks to extract information from bar charts and pie charts, but they also adopted some extra low-level operations to match the textual and numeric information.

In this paper, we propose a new reverse-engineering method for bar charts (sketched in Figure 1). The main idea is to use neural networks for textual and numeric information extraction. For textual information, we use Faster-RCNN, which is a neural network-based object detection model, to simultaneously localize and classify textual information. This approach improves efficiency by simplifying the flow of prior work and reducing the low-level operations. For numeric information, we use an encoder–decoder framework that integrates convolutional neural network (CNN) and recurrent neural network (RNN). This framework can learn to directly transform chart images into numeric values without rules and thus achieves a good accuracy and robustness. This framework presents a potential generalization for other types of charts because it first understands chart images and then extracts numeric information by order without additional neural networks for matching the textual and numeric information[14]. An attention mechanism, which is usually a shallow feed-forward neural network, is included to further increase accuracy. It helps the encoder–decoder framework produce highly accurate results by properly distributing "visual attention" over the chart image.

A synthetic dataset (30,300 bar charts) and a real-world dataset (180 bar charts) are used to train and evaluate our method. For textual information localization and classification, we achieve F1 scores of 0.89 and 0.80 for synthetic and real-world data, respectively. For numeric information, we achieve accuracies of 0.91 and 0.78 for synthetic and real-world data, respectively (under a criterion of 5% deviation). The entire extraction process for one bar chart costs approximately 3 seconds on average on a machine with local OCR engine (2 Xeon CPUs, 12 Gi RAM, and 1 NVIDIA Tesla K80 GPU). In summary, this work presents a new reverse-engineering method for bar charts. Our method improves the efficiency of textual information extraction by using a neural network-based object detection model. An encoder–decoder framework with attention mechanism is used to achieve highly robust and accurate numeric information extraction.

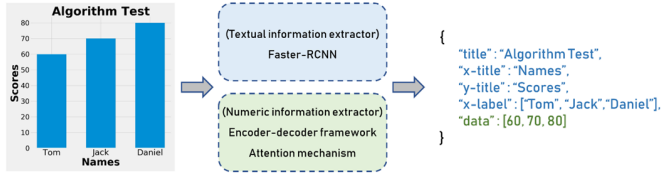

Figure 1. Reverse-engineering bar chart. Given the visual representation of a bar chart, we extract its underlying information by using two components. The first component uses object detection model for textual information extraction. The second component uses an encoder–decoder framework with attention mechanism to extract numeric information.

## 2 RELATED WORK

Our work builds on two areas of related work: textual and numeric information extraction from charts.

### 2.1 Textual Information Extraction

Textual information extraction usually follows a bottom-up approach, from pixel to word, and includes three steps, namely, localizing the text elements, recognizing them for the underlying text, and classifying them by text types. Many previous works[2, 6, 15-17] rely on user interaction to obtain the texts or simply assume that the textual information is given beforehand because effectively extracting the textual information is a difficult task. For text localization, many works use the connected component to merge pixels into words. Huang et al.[18] exploit the connected component analysis to separate textual and numeric information in charts. Jayant et al.[19] and Boschen et al.[20] take an additional step to infer text orientation, which is needed in text recognition. For text recognition, out-of-the-box tools, such as Microsoft OCR[21] and Tesseract[22], are usually used in practice. It is possible to use general-purpose OCR engine for text localization and recognition. However, two experimental studies [6] [7] show that text localization with OCR performs poorly on chart images. For text classification, most works use geometric features of text because different text types vary in their placement. Huang et al.[18] investigate text classification by using geometric relationships between text and graphical elements. Chen et al.[5] use text bounding boxes to classify text in DiagramFlyer. In addition to bounding boxes, Choudhury et al.[17] include text content for classification.

Poco et al.[7] and Dai et al.[8] utilized neural networks to extract textual information. They first used the neural networks to binarize an image, then removed non-text pixels, and finally applied the connected components algorithm to locate and distinguish the text objects. Despite acceptable results, their methods contain multiple low-level operations and have to obey many fixed rules, such as constructing artificial features by geometric information for text classification. In this paper, we use a neural network-based object detection method for textual information extraction, which can simplify the extraction process and enhance the robustness.

### 2.2 Numeric Information Extraction

The prior works on numeric information extraction take two main classes of approaches: user interaction and image processing techniques. ChartSense[15] is an interactive chart data extraction system that uses user interactions to mark key graphical elements. Mendez et al.[16] introduce iVoLVER, which is a system that supports interactive data acquisition from multiple source types (including chart images). Data Thief[23] and WebPlotDigitizer[24] are two publicly available systems that use user interaction to extract information from charts. Although interactive systems can achieve high accuracy by exploiting user intelligence, they usually suffer from involving users in an exhausting and time-consuming extraction process. Users may also have to learn the various tools that these systems offer prior to information extraction.

Digital image processing is a well-studied field and provides numerous useful methods to handle image-related problems. Many prior works use these methods and develop task-specific rules to automatically extract chart information. Zhou et al.[25] use boundary tracing and Hough transform to identify bars in bar charts. Huang et al.[18] use edge maps and rules to extract graphical elements from several chart types. The two works only extract the graphical elements from charts without recovering the underlying numeric information. ReVision[2], Dai et al.[8], and Al-Zaidy et al.[26, 27] follow a similar approach to extract numeric information. First, the bar chart images are preprocessed, such as thresholding and color space conversion. Then, the connected component analysis is used to merge pixels that form rectangular shapes. These shapes are filtered to remove background noise according to fixed rules, such as fill rate and color difference. Finally, the heights of shapes are mapped back to numeric values. These techniques usually have difficulty achieving high accuracy because of the limited capabilities of traditional image processing techniques and the disadvantage (i.e., lack of robustness) of using fixed extraction rules. The successful development of rules also requires a thorough understanding of the charts, such as the encoding of data. In this work, we use neural networks to extract numeric information. Neural networks feature self-learning ability, which eliminates the need for rules, and have better performance in many image-related problems than traditional image processing techniques.

## 3 PRELIMINARY

In this section, we describe the two datasets that we use to train and evaluate our neural network models. We also present the design consideration of our method.

### 3.1 Bar Chart Assumption

Bar charts have many different forms and design styles[28-39]. To scope the research space, we have the following simplifying assumptions: (1) only one bar chart per image is considered; (2) bar charts do not contain 3D effects; (3) elements in bar charts do not overlap each other; (4) bar charts contain only horizontally and vertically oriented texts; (5) a coordinate with numeric axis labels is present in the bar charts; (6) bar charts do not contain stacked bars; (7) bar charts contain only horizontally or vertically oriented bars. Under these assumptions, the bar charts still have many variations. Figure 2 shows a few samples that satisfy the assumptions.

### 3.2 Data collection and generation

Neural networks have an inherent requirement for labeled data[40]. In general, more labeled data gives better performance. However, obtaining a large amount of labeled data can be a difficult task. After collecting and studying some real-world bar chart visualizations, we find that bar charts are similar to each other if we ignore the design style, color schemes, and other irrelevant factors. Therefore, we propose to first train neural networks with a large amount of synthetic data and then fine-tune the networks with a small amount of real-world data.

The real-world data is collected through search engines. We collect 180 images (60 images per search engine) from three popular search engines: Bing, Google, and Yahoo. We use the key word "bar chart" in these search engines. We eliminate the charts that do not satisfy the assumptions during collection, and duplicate charts from different search engines are avoided by comparing charts before placing them into the collected set. These chart visualizations are then labeled manually to generate the ground-truth information. Finally, we randomly shuffle and split these charts into training and test sets (Table 1). Figure 2 (a) shows sample bar charts from the collected set.

The synthetic data is generated by a Python script with the Matplotlib module[41], which provides many useful and flexible functionalities to generate publication quality charts. Using this method, we can generate an arbitrarily large dataset that systematically varies the visual encodings. We can easily obtain the underlying data, such as bounding boxes in pixel coordinate and bar heights, because the dataset is generated programmatically. We first investigate the variation in real-world chart visualizations to synthesize data as realistic as possible. Table 2 lists the various factors that we consider when generating synthetic bar charts. The investigation later guides us in the design of synthetic data. We generate a dataset of 30,300 bar chart visualizations in this work. The dataset is then randomly shuffled and split into training and test sets (Table 1).

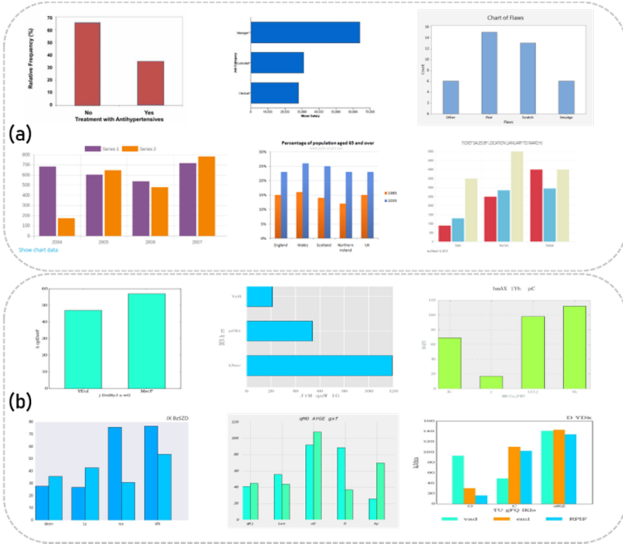

Figure 2 Various visual representations of bar charts. The bar charts in (a) are collected from search engines. The bar charts in (b) are generated by Python script. The synthetic bar charts are varied systematically to approximate the real-world bar charts.

Table 1. Dataset statistics

| Dataset | Training | Test | Total |
| --- | --- | --- | --- |
| Real-world dataset | 150 | 30 | 180 |
| Synthetic dataset | 30000 | 300 | 30300 |

Table 2. Considerations in generating synthetic bar charts

| Factor | Specification |
| --- | --- |
| Design style | 25 styles |
| Figure size | 300 to 900 pixels |
| Font and font size | 5 fonts and 7 relative font sizes |
| Text length | 1 to 15 characters |
| Title position | top left, top center, top right |
| Legend position | top center, upper right, center right, and bottom center |
| Bar orientation | horizontal and vertical |
| Bar series | 2 to 5 |
| Bars per series | 1 to 3 |
| Bar width and height | randomly chosen according to figure size |
| Bar color | randomly chosen according to color maps |

In a synthetic bar chart generation process, the workflow can be described as follows. First, an image size with varying aspect ratio is randomly determined. Then, we randomly choose one design style from many candidate ones to change the overall look-and-feel of the synthetic charts. Font and font size are randomly determined for the texts. We note that all text types use the same font but may use different font sizes in a single chart. The texts are randomly generated and have varying lengths. We then randomly decide the inclusion of a title. If it is included, then a position is randomly chosen for it. Legend is processed in a similar manner to title, but the position of legend has more choices. The inclusion of axis title is randomly decided, and x-axis and y-axis use the same configuration except for the texts. We assume that axis labels always exist, as in nearly all real-world charts.

The numeric data is also generated randomly. In the case of bar charts, we need to randomly choose the number of series and the number of bars per series and determine the height of every bar. Bar orientation is randomly chosen from two values, namely, horizontal and vertical. We also need to choose colors for these bars from specific color maps. Figure 2 (b) presents bar chart samples generated using the above-mentioned workflow.

### 3.3 Design Consideration

Our method aims to effectively extract textual and numeric information from bar charts. In this sub-section, we provide design consideration regarding the techniques that are used for the two types of information extractions, respectively.

For textual information extraction, the problem of the prior works is inefficiency due to the use of many low-level operations and slightly excessive procedures. We propose to use object detection models to address this problem. First, object detection models can localize and classify textual information simultaneously, which reduces the extra procedures. Second, object detection models do not involve such low-level operations as pixel classification and merge and can be accelerated using optimized neural network structures. Object detection has been extensively investigated in recent years, and various models have been proposed. These models can generally be organized into two main categories[10]: one- and two-stage frameworks. One-stage framework directly performs classification and bounding box regression. The two-stage framework includes an additional region proposal stage, which usually has better results than the one-stage one. Although the one-stage framework is generally faster and has competitive result as the two-stage framework, it does not perform well in detecting small objects. Considering that text elements in bar chart visualizations are usually of small size, we decide to use a two-stage framework. Specifically, we use Faster-RCNN[43], which is a state-of-the-art object detection model and can be used in real-time scenarios. The advantage makes it suitable for improving efficiency in our work.

For numeric information extraction, the prior works fail to achieve high accuracy. We consider a two-step approach to achieve high extraction accuracy. As the first step, we propose to use an encoder–decoder framework to extract numeric information. As the first step, we propose to use an encoder–decoder framework to extract numeric information because using object detection models is difficult to be applied to other types of charts. The inputs (bar chart images) and outputs (numeric values of bar height) have different forms. The encoder–decoder framework plays the role of a "translator" that first understands the chart images and then translates them into numeric values. In this framework, the encoder deals with chart images and extracts the key features from them; the decoder works with numeric values and interprets the features to produce desired data. We implement the encoder with CNN in this study, which can effectively extract features from images. CNN identifies the bars and their properties and produces a concise feature map. We use RNN, which is designed for processing and generating sequence data, to implement the decoder. RNN consumes the feature map and iteratively generates a description for each bar in the chart. We assume that the bars within a bar chart form a sequence, from top to bottom or from left to right. This assumption eases the extraction process and post-processing.

Although the encoder–decoder framework can extract numeric information, preliminary experiments show that the results are less accurate. We also find that it generates fewer or more bars than the ground-truth bars in some cases. As the second step, we introduce an attention mechanism, which is used in neural machine translation[42], into the framework to solve these issues. Intuitively, it helps neural network focus more on the words for translation and less on the other context words, which simulates the translation process of human to generate accurate translation. Xu et al.[11] introduce this mechanism into the vision field and use an attention-based model to generate accurate image caption, which achieves state-of-the-art performance. Their work mainly involves an iterative image-to-sequence process, which is analogous to numeric information extraction from bar charts. In this work, the attention mechanism helps the encoder–decoder framework focus on the bar it extracts to ensure highly accurate numeric value. It also helps the framework determine the correct number of bars in the chart, as shown by experiments (Section 5.2).

## 4 EXTRACTION METHOD

Our method mainly consists of two components: textual and numeric information extractions. In this section, we first detail the two components. Then, we describe the training process. Finally, we present the data recovery procedure (i.e., post-processing).

### 4.1 Extracting Textual Information

In textual information extraction, we use Faster-RCNN to perform concurrent localization and classification of textual elements. Faster-RCNN mainly consists of a feature extractor, a region proposal network (RPN), a multi-class classifier, and a bounding box regressor, which are all neural networks. The feature extractor takes an image as input and produces a feature map. The RPN generates candidate regions containing interesting object from the feature map. The candidate regions are then inputted into the classifier and regressor to obtain the classes and bounding boxes. In this work, the classifier outputs vectors of seven elements for each candidate region. Each element of the vector corresponds to a probability that the object in the candidate region belongs to one of the six roles or non-text: $\{p_1, p_2, ..., p_7\}$. The regressor outputs bounding boxes for each candidate region. The bounding box consists of the center coordinates and the width and height of the box: $\{t_x, t_y, t_w, t_h\}$.

Figure 3 summarizes our pipeline for textual information extraction, which involves three steps: localizing and classifying textual elements, obtaining sub-images of individual textual elements, and applying OCR and sorting (steps (1) to (3) in Figure 3). First, input bar chart image (Figure 3 (a)) into Faster-RCNN that performs concurrent localization and classification of textual elements. Figure 3 (b) shows the bar chart image after the bounding boxes and text classes produced by Faster-RCNN are applied. The bounding boxes are then used to crop the bar chart image, thereby yielding the sub-images in Figure 3 (c). OCR engine is applied on these sub-images to recover the underlying texts. To ease the matching between textual and numeric information, the axis labels and legend, if exist, are sorted according to their bounding box coordinates and alignment. Finally, the texts and their classes are combined to produce the result of textual information extraction (Figure 3 (d)).

We present some details in our extraction process. For each detected textual element, the output of the Faster-RCNN also includes a confidence. We set a threshold of 0.9 to accept only detections with high confidence, which suppresses background noise. The sub-images are first converted into binary images[44] and scaled by a factor of 2 before OCR recognition. For each sub-image, OCR engine is applied three times[7]: sub-image rotated 0°, sub-image rotated 90° clockwise, and sub-image rotated 90° counter-clockwise. The rotations are conducted to handle the text orientation problem. The output of OCR consists of the text string and a confidence score. Among the three recognitions, the one with the highest score is selected as the recognition result for the sub-image.

### 4.2 Extracting Numeric Information

We use three neural network models, namely, the encoder, the decoder, and the attention mechanism, to extract numeric information. The entire extraction pipeline for numeric information, which contains five steps, is presented in Figure 4. The encoder takes bar chart image as input and performs various operations to extract features, such as two-dimensional convolution and down sampling. It then produces a sequence of feature vectors (Figure 4 (a)). The attention model takes the feature vectors and the hidden state (Figure 4 (f)) of the decoder from the previous iteration and generates an attention vector (Figure 4 (b)) for the current iteration. The attention vector contains information about the generation process (e.g., information about which bar to focus on in the current iteration) due to the use of hidden state. The attention vector and feature vectors are then combined (Figure 4 (c)) and concatenated with the bar vector (Figure 4 (e)) produced by the decoder from the previous iteration to generate the context vector (Figure 4 (d)). With the context vector, the decoder produces bar vector for the currently focused bar in the chart. For emphasis, the decoder works in an iterative manner that produces information about one bar in each iteration. The attention model and the decoder continue the above-mentioned process (steps (2) to (5) in Figure 4) until all the bars in the chart are extracted.

We note a difference between the training and evaluation processes. As shown in Figure 4, the bar vector in step (4) during the evaluation process is generated by the decoder from the previous iteration. In the training process, this bar vector is the ground-truth bar vector that precedes the currently focused bar vector. The bar information is trained to be generated from the leftmost bar to the right in vertical-horizontal-oriented bar charts. This design eases the matching.

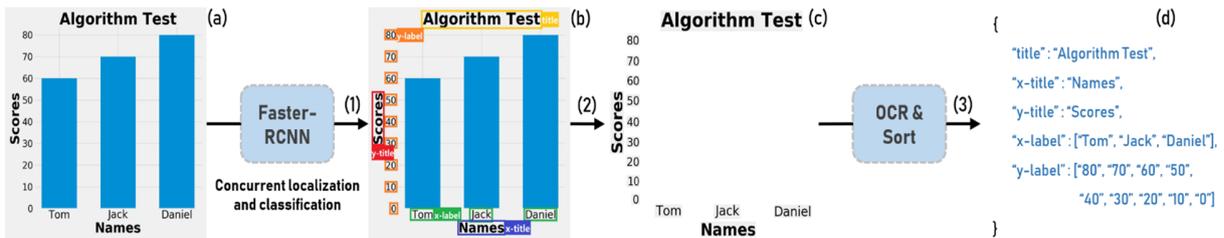

Figure 3. Pipeline for extracting textual information. (a) Input bar chart image. (b) Bar chart image with textual elements localized and classified. (c) Sub-images obtained using the bounding boxes of (b). (d) Resulting textual information. Steps (1) to (3) constitute the entire pipeline.

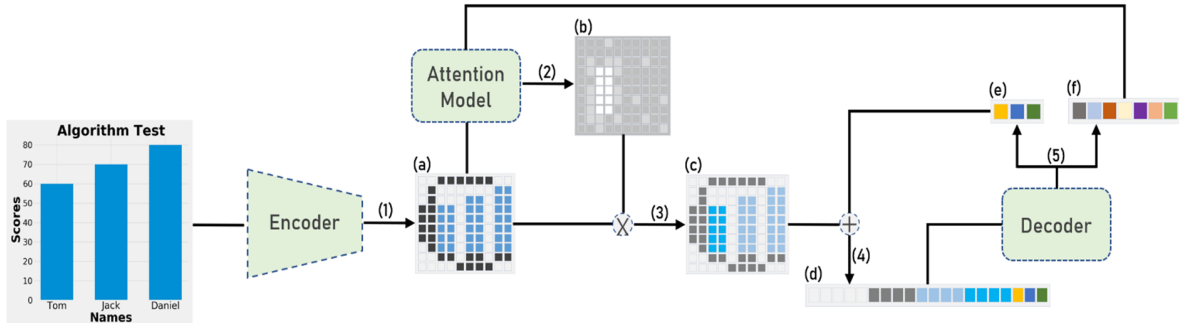

Figure 4. Pipeline for extracting numeric information. (a) Feature vector. (b) Attention vector. (c) Feature–attention vector. (d) Context vector. (e) Bar vector. (f) Hidden state. Step (1) produces the feature vector. Steps (2) to (5) iteratively generate the numeric information, that is, bar vector.

### 4.2.1 Encoder

The encoder is responsible for understanding the bar chart images and extracting features from them. It is mainly composed of CNN that has a layered structure to gradually learn the features. In general, deeper CNNs have great abilities to learn the desired features. The disadvantage is that deep CNNs may be hard to train, have a large size, and run slowly. When determining the encoder structure, we need to achieve a balance. In this work, we use the Xception[45] as the base of our encoder. Xception is a well-designed CNN structure that considers performance and cost. It uses depthwise separable convolution to achieve high performance while keeping light weight.

Xception, which is a relatively deep network, has 126 layers. Nearly 30% of the layers are convolutional layers that constitute the feature extraction base of the network[45]. Because deep networks are hard to train (due to vanishing and/or exploding gradients), Xception uses residual connections[46] to help address the issue. Xception also includes layers that are commonly found in CNNs: dropout layers, max pooling layers, and rectified linear unit activation layers. Xception takes images of size $299 \times 299$ as input. The output of the last convolutional layer has a shape of $10 \times 10 \times 2048$. The first two dimensions still keep the spatial correspondence with the original image to easily apply the attention mechanism.

In addition to Xception, the encoder includes two layers. A reshape layer first transforms the output of the last convolutional layer of Xception into shape $100 \times 2048$. Then, a fully connected layer reduces the last dimension, thereby yielding the final output as a sequence of 100 feature vectors of size 256: $F = \{f_1, f_2, ..., f_{100}\}, f_i \in R^{256}$. The two layers are added for the attention model. Reshape is needed because our attention model takes a two-dimensional vector as input (excluding the batch size). Reducing the last dimension limits the input size of the attention model and thus limits its parameter space for easy training. Note that the fully connected layer is applied only along the last dimension. This design preserves the correspondence between the feature vectors and the original image. With the correspondence, attention model can highlight a subset of the feature vectors that correspond to parts of the image that the decoder should focus on. This design also allows us to investigate the attention model by combining the output of the attention model and the original image, as shown in Section 5.2.

### 4.2.2 Attention Model

The attention model is the key to achieving high accuracy when reverse-engineering bar charts. Attention model usually comprises shallow neural network or multi-layer perceptron. We keep the attention model simple because learning to generate visual attention of bar charts should be relatively easy. In this study, we use a two-layer neural network to implement the attention model.

The attention model takes as input the feature vectors $F$ and the hidden state $h_{t-1}$ generated by the decoder in the $t-1$ iteration and outputs an attention vector $\alpha_t$ for the current iteration. The model can be described by the following equations:

$$e_t = \theta(F, h_{t-1}), \alpha_t = softmax(e_t),$$

where $\theta(\cdot)$ is a layer that first transforms $F$ and $h_{t-1}$, respectively, and then merges the two resulting vectors by addition and non-linear transformation. $e_t$ is then passed to the next layer containing softmax activation function. The output of the attention model is a sequence of scalars, namely, the attention vector in this work. The attention vector has the same length as $F$, and each element corresponds to a feature vector, which further corresponds to a portion of the original image. The elements in the attention vector can be interpreted as the relative importance that should be given to the portions of the image.

Two ways can be used to apply the attention vector: stochastic attention and deterministic attention[11]. For simplicity, we use deterministic attention that can be expressed as

$$\widehat{f}_t = \sum_{i=1}^{100} \alpha_{ti} f_i, \alpha_{ti} \in \alpha_t,$$

where $\widehat{f}_t$ is called the feature–attention vector. This vector is used in the decoder to guide the generation of bar information. Note that the attention model also works in an iterative manner. It takes the hidden state of the decoder from the previous iteration and outputs the attention vector for the current iteration that informs the decoder where to "look" next.

### 4.2.3 Decoder

The decoder interprets the context vector and generates the numeric information iteratively. The output for each bar has the following form: $b_t = \{x_t, y_t, v_t\}$, where $x_t$ and $y_t$ are the center coordinates of the bar and $v_t$ is its normalized height. The center coordinates are included to help the decoder generate the bar information sequentially. Because the bar information is generated as a sequence: $B = \{b_1, b_2, ..., b_n\}$, we use RNN as the base of the decoder. Specifically, long short-term memory (LSTM)[47] is selected. LSTM can avoid the problem of vanishing and (or) exploding gradients that is common in traditional RNNs. It also has the benefit of memorizing long-term dependencies, which is useful when dealing with sequence data.

We use only one LSTM in the decoder because the bar sequences are usually not very long, and the bar vectors have a simple form. LSTM can be described mathematically by the following equations:

$$i_t = \sigma(W_i[x_t, h_{t-1}] + \beta_i),$$
$$r_t = \sigma(W_f[x_t, h_{t-1}] + \beta_f),$$
$$o_t = \sigma(W_o[x_t, h_{t-1}] + \beta_o),$$
$$c_t = r_t * c_{t-1} + i_t * \emptyset(W_c[x_t, h_{t-1}] + \beta_c),$$
$$h_t = o_t * \emptyset(c_t),$$

where $W$s are the weight matrices, and $\beta$s are the biases. $x_t$ is the input vector at iteration $t$ (that is, the context vector in this work), $h_{t-1}$ is the hidden state from the previous iteration, and $c_{t-1}$ is the cell state from the previous iteration. $r_t$ is the output of forget gate, $i_t$ is the output of input gate, and $o_t$ is the output of the output gate. $\sigma(\cdot)$ and $\emptyset(\cdot)$ are the sigmoid and hyperbolic tangent activation functions, respectively. The hidden state and cell state carry information from one iteration to another. The different gates control how information flows through LSTM. In this work, the LSTM takes as input the concatenation of the feature–attention vector $\widehat{f}_t$ and the bar vector $b_{t-1}$ from the previous iteration. It then outputs the hidden state vector $h_t$ of size 512 (the number of units in our LSTM). Following the LSTM is a fully connected layer, which transforms the output of the LSTM into the bar vector $b_t$ of the current iteration.

### 4.3 Training

The training process can be divided into two parts, which can proceed concurrently. The first part trains the object detection model, and the second part trains the encoder–decoder framework with attention mechanism. Both trainings follow the same pattern: We first use 30,000 synthetic bar charts to train the model, so that the neural network can learn the required parameters; and then use 150 real-world bar charts to fine-tune and enhance the generalization of the neural network. The training data is shown in Table 1. The training process is performed on a system with 2 Xeon CPUs, 12 Gi RAM, and 1 NVIDIA Tesla K80 GPU.

The training of Faster-RCNN is well documented in its paper[43]. We only present specific points here. We use the parameters pre-trained on the COCO dataset[48] as a starting point to reduce training time. The training data consists of chart images and corresponding labels. The label includes bounding box coordinates and text roles (i.e., classes). A learning rate of 0.0003 is used for 300 epochs and 0.00003 for 50 epochs (fine-tuning).

The training data for our encoder–decoder framework with attention mechanism contains chart images and bar vector sequences. We use the special vectors {1,1,1} and {0,0,0} as the start and end bar vector of the sequence, respectively. Because the numbers of bars vary from chart to chart, we preprocess the bar sequences by padding them with the end vector to the same maximum length. Mean squared error (MSE) is used as the loss function. We use the Adam optimizer with a learning rate of 0.001 to update the parameters of the models for 300 epochs, 0.0001 for 100 epochs, and 0.00001 for 50 epochs (fine-tuning).

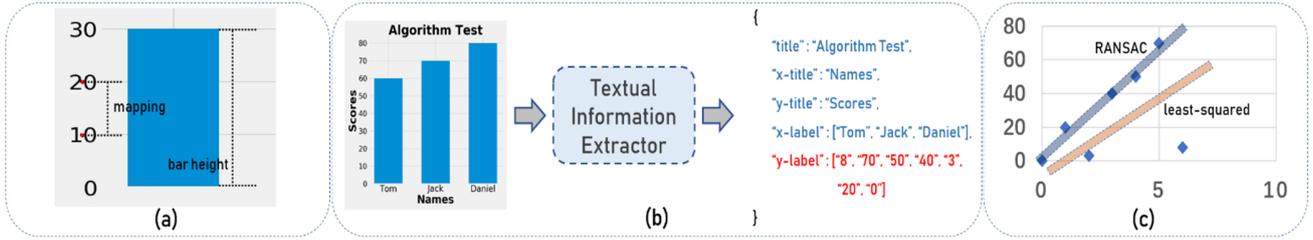

Figure 5. Robust data recovery. (a) Relationship between mapping and bar height. (b) Errors occur when extracting textual information. (c) RANSAC regression is more robust to outliers than least-squared linear regression.

### 4.4 Recovering Data

The bar vectors, which contain the center coordinates of bars and normalized bar heights, are floating-point numbers in the range of 0 to 1. These numbers are measured using the top-left corner of the image as origin and the image width and height as the unit length of the x-axis and y-axis, respectively (we call this coordinate the normalized pixel coordinate). To recover the original bar heights, we need a mapping (Figure 5 (a)) between the normalized pixel and chart coordinates (the coordinate drawn on the image). We use the y-axis labels and their coordinates to calculate the mapping (we always call the axis containing the numeric axis labels the y-axis). Suppose we have extracted a set of y-axis labels:

$$y_i = \{l_i, t_{xi}, t_{yi}\}, i = 1,2,3,...,$$

where $l_i$ is the numeric value of the label, and $(t_{xi}, t_{yi})$ is the center coordinate of the label. Because the coordinate $(t_{xi}, t_{yi})$ may contain localization and (or) classification errors (other text types are misclassified as y-axis labels), we need to first filter these errors. We observe that y-axis labels are nearly always placed equal-spaced, and outliers (those labels with errors) are less common than inliers. Thus, we can use regression methods to filter outliers. RANSAC regression[49] is chosen due to its robustness to outliers compared with other regression methods, such as least-squared linear regression. $l_i$ is filtered similarly due to OCR recognition errors. We use only those y-axis labels that remain after the two filtering. Figure 5 (b) and (c) show an example for filtering $l_i$, where (b) shows the localization and recognition errors (text in red) during textual information extraction, and (c) shows that we can keep those $l_i$s that are close to the RANSAC line to filter outliers.

To calculate the mapping, we also need to determine the orientation of the bars. The orientation can be deduced by calculating the variance of the x and y coordinates of the y-axis labels, respectively. If the variance of the x coordinates is greater than that of the y coordinates ($var(\cdot)$ is the function that calculates variance):

$$var(t_{x1}, t_{x2}, t_{x3}, ...) > var(t_{y1}, t_{y2}, t_{y3}, ...),$$

then the orientation of the bars is horizontal. Otherwise, the orientation is vertical. The mapping can be achieved as follows. We first calculate the difference between a pair of y-axis labels ($abs(\cdot)$ is the function that calculates absolute value):

$$abs(l_i - l_j).$$

We then compute the difference between their coordinates. If the orientation is horizontal, then we use the x coordinate in the calculation; otherwise, we use the y coordinate:

$$abs(t_{xi} - t_{xj}) \text{ or } abs(t_{yi} - t_{yj}).$$

The mapping is computed as：

$$\frac{abs(l_i - l_j)}{abs(t_{xi} - t_{xj})},$$

if the orientation is horizontal (similarly for vertical). To reduce noise, we randomly choose five pairs of y-axis labels, if possible, and use the average of their respective mapping as the final mapping. Finally, the bar height in the chart coordinate can be recovered by scaling the normalized bar heights by the mapping.

## 5 EXPERIMENTS

In this section, we demonstrate the effectiveness of our method using two datasets: the synthetic and real-world datasets. We also show how attention mechanism assists the encoder–decoder framework in improving the extraction accuracy.

### 5.1 Quantitative Analysis

We perform quantitative analyses for textual and numeric information extractions separately. The experiments are run on the same machine as training.

#### 5.1.1 Textual information extraction analysis

The evaluation of the textual information extractor is reported in this sub-section. Because OCR is not our concern, we do not evaluate the performance of character recognition. We show two popular object detection metrics for this evaluation because we use the idea of object detection to extract textual information. Specifically, the COCO[47] and Pascal VOC[50] metrics are used.

Table 3. Text detection result using the COCO metric.

| Dataset | AP@[.5:.95] | AR@[.5:.95] |
|---|---|---|
| Synthetic data | 0.889 | 0.893 |
| Real-world data | 0.796 | 0.802 |

Table 4. Text detection result using the Pascal VOC metric.

| Dataset | Synthetic data | Real-world data |
|---|---|---|
| title | 1.000 | 0.896 |
| legend | 1.000 | 0.921 |
| x-axis title | 1.000 | 0.889 |
| y-axis title | 1.000 | 0.884 |
| x-axis label | 0.985 | 0.865 |
| y-axis label | 0.983 | 0.862 |

Table 3 shows the result of the COCO metric. The average precision (AP) and average recall (AR) are averaged over 10 intersection over unions (IOU) from 0.5 to 0.95 with a step of 0.05 and over all six text classes. We achieve relatively good results for the synthetic and real-world datasets. The performance difference between the two datasets may be attributed to the enormously various real-world charts and our fixed-pattern synthetic charts. Table 4 shows the result of the Pascal VOC metric. The AP is based on an IOU of 0.5. High precisions are achieved for both datasets. We note that the precisions for x- and y-axis labels are lower than those of the other text classes. This finding may be due to the fact that the numbers of x and y labels are relatively large and their placements are similar, which leads to confusion.

Dai et al.[8] reported an average precision of 0.82 for text role classification on 59 real-world bar charts collected from search engines. Our method achieves an average precision of 0.88 for text role classification on the real-world test set. Although our test set is not completely same with that of [8], this result indicates that our method outperforms in the textual information extraction to a certain extent. On the experiment machine with local OCR, our method takes an average of only 2.4 seconds to perform the entire textual information extraction for one bar chart image.

### 5.1.2 Numeric information extraction analysis

We conduct two groups of experiments, one for each test sets. Each group has two experiments, one for OCR without correction and one for OCR with correction. Although OCR is not the focus of this work, it affects the extraction result. During data recovery, the character recognition accuracy of OCR affects the correctness of the mapping despite the use of RANSAC regression. To show the result after eliminating this irrelevant factor, the experiments, OCR with correction, are performed. These experiments are the same as OCR without correction except that we use the ground-truth data to correct OCR recognition errors. For instance, if the OCR recognizes a ground-truth y label "80" as "8," then we correct the error and yield "80."

We use accuracy to rate the extraction results. Specifically, the following criterion is used[8]:

$$\frac{abs(h_g - h_p)}{h_g} \leq \varepsilon,$$

where $h_g$ is the ground-truth bar height, $h_p$ is the predicted bar height, and $\varepsilon$ is a threshold ($0 \leq \varepsilon$). $\varepsilon$ controls the strictness of the evaluation. A small $\varepsilon$ indicates a strict evaluation. For a single bar chart, if the model predicts more (or fewer) bars than ground-truth bars, we first pad the shorter one to the same length as the longer one with pre-defined maximum bar heights (the maximum bar height always results in invalid criterion). Then, we compare the two bar sequences. If a predicted bar height and its corresponding ground-truth bar height meet the criterion, we consider that the bar is correctly extracted. And we define the accuracy rate of numeric information extraction:

$$acc = \frac{N_{correct}}{N_{total}},$$

where $N_{correct}$ is the total number of correctly extracted bars in the test set, and $N_{total}$ is the total number of bars in the test set after padded.

Table 5. Numeric information extraction accuracy, $\varepsilon = 0.05$.

| Dataset | Accuracy (with OCR correction) | Accuracy (without OCR correction) |
| --- | --- | --- |
| Synthetic data | 91% | 82% |
| Real-world data | 78% | 71% |

Table 6. Numeric information extraction accuracy, $\varepsilon = 0.02$.

| Dataset | Accuracy(with OCR correction) | Accuracy(without OCR correction) |
| --- | --- | --- |
| Synthetic data | 86% | 79% |
| Real-world data | 71% | 67% |

Tables 5 and 6 show the extraction results for $\varepsilon = 0.05$ and $\varepsilon = 0.02$, respectively. In Table 5, our method correctly extracts 91% of the numeric information for the synthetic dataset with OCR correction. For the synthetic dataset without OCR correction, the accuracy reduces to 82%. The reduction indicates the significant influence of the OCR recognition. A total of 78% and 71% of the numeric information are correctly extracted for the real-world dataset with and without OCR correction, respectively. The decreased accuracies indicate that a difference exists between the real-world and synthetic bar charts. This finding is not surprising because we use a large amount of synthetic data to train the models. We believe that our method can achieve comparable accuracy if sufficient labeled real-world training data is accessible. Table 6 shows a similar result to Table 5 but with decreased accuracies due to the use of a stricter $\varepsilon$. Our numeric information extraction accurately predicts the number of the bars in 99% of the synthetic bar charts and 92% of the true real-world bar charts, which are better than the results (74% of the bar charts on a dataset of 59 real-world bar charts) reported in [8]. And our results are also better than the results reported in [14] (79.4% of the bar charts on a dataset 0f 3000 simulated bar charts witch which is mostly the same generation of our synthetic dataset). These values are calculated by the formula: $B_{correct}/B_{total}$, where $B_{correct}$ is the number of bar charts with the correct number of bars predictions, and $B_{total}$ is the total number of the bar charts. Using a local Tesseract OCR engine, our model uses approximately 3 seconds on average to extract textual and numeric information for a single bar chart.

Figure 6 shows some typical extraction results (without OCR correction). For textual information extraction, most textual elements are correctly localized and classified. Our method properly handles many different situations, such as different title positions (Figure 6 (b), (c), and (d)), titles and axis titles with white spaces (Figure 6 (j) and (h)), and different legend positions (Figure 6 (e), (f), (h), and (l)). For the three sample real-world bar charts (Figure 6 (m) to (o)), we correctly extract all the textual information. We note that our method fails some detection: the second x-label in Figure 6 (h) and the third x-label in Figure 6 (l). The failure may be due to the small sizes of these single-character x-labels. Most prediction strings do not exactly match the ground-truth strings because of OCR recognition errors (e.g., OCR fails to recognize the last single-character x-label in Figure 6 (l)). For numeric information extraction, our method achieves good accuracy and robustness. It properly handles horizontal and vertical bar charts (e.g., (b), (c), (m), and (n) in Figure 6) and many different design styles and backgrounds. The synthetic dataset contains twelve bar combinations (2 to 5 bar series and 1 to 3 bars per series), as shown in Figure 6 (a) to (l). For all these combinations, most numeric information is correctly extracted: all bars are correctly identified, and the deviation is minor. Our method works especially well on bar charts with fewer bars, such as Figure 6 (a) and (b). As the number of bars increases, the performance may degrade slightly, such as Figure 6 (l). The degradation may be due to the fact that the bars in dense bar charts are too thin to handle. For the three real-world bar charts, the numeric information extractor performs reasonably. Despite OCR recognition errors, we still achieve robust data recovery because of RANSAC regression. Additional extraction results are provided in the appendix.

### 5.2 Effectiveness of attention mechanism

In this paper, the attention mechanism helps the encoder-decoder framework achieve a highly robust and accurate numerical information extraction. We conduct three groups of numerical extraction experiments to illustrate the effectiveness of this mechanism. We use the synthetic dataset with OCR correction for the experiments and the same accuracy criterion in Section 5.1.2. The first group uses the encoder-decoder framework with attention mechanism (our proposed method); the second group is similar to the first one, but its attention model does not update the parameters during the training process; the third group uses an encoder-decoder framework without attention mechanism. The models of these three groups are trained with the same configuration, such as the same training set, training rounds, parameter initialization method, etc.

Table 7. Comparison of trainable and non-trainable attention models.

| Setup | Accuracy |
| --- | --- |
| Trainable attention model | 91% |
| Non-trainable attention model | 84% |
| No attention model | 85% |

Table 7 shows the extraction results for $\varepsilon = 0.05$. In the first group, our method correctly extracts 91% of the numeric information for the synthetic dataset with OCR correction, as shown in Table 5. However, the accuracy reduces to 84% and 85% in the second and the third group, respectively. The results indicate that the attention mechanism can improve the accuracy of numeric information extraction. The second group uses an untrainable attention mechanism, but its accuracy is 1% lower than the third group (without the attention mechanism). We speculate that this is caused by a potential information loss. The parameters of the attention mechanism for the second group are randomly initialized and are not learned during the training process, and the weight vectors that it generates are equivalent to random noises.

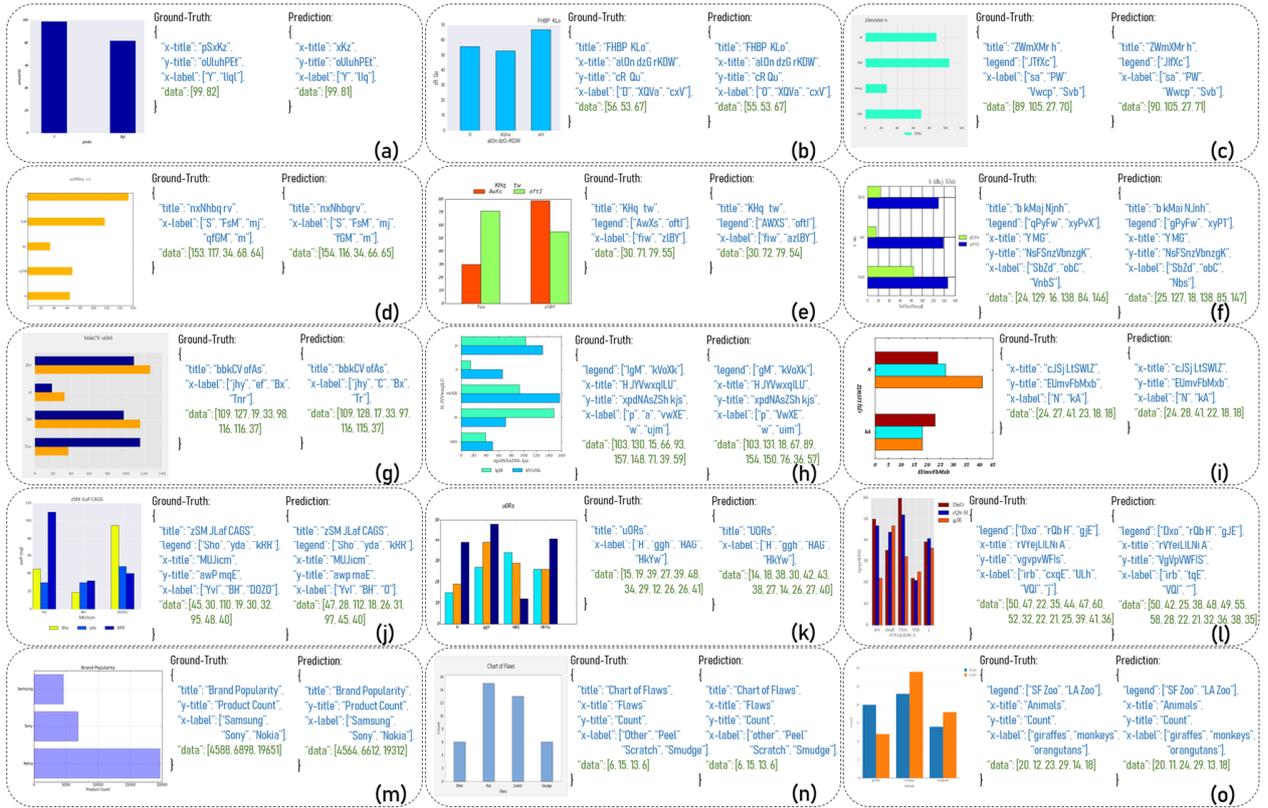

Figure 6. Some typical extraction results. (a) to (l) are the synthetic bar charts. (m) to (o) are the real-world bar charts. These bar charts are representatives of our two datasets.

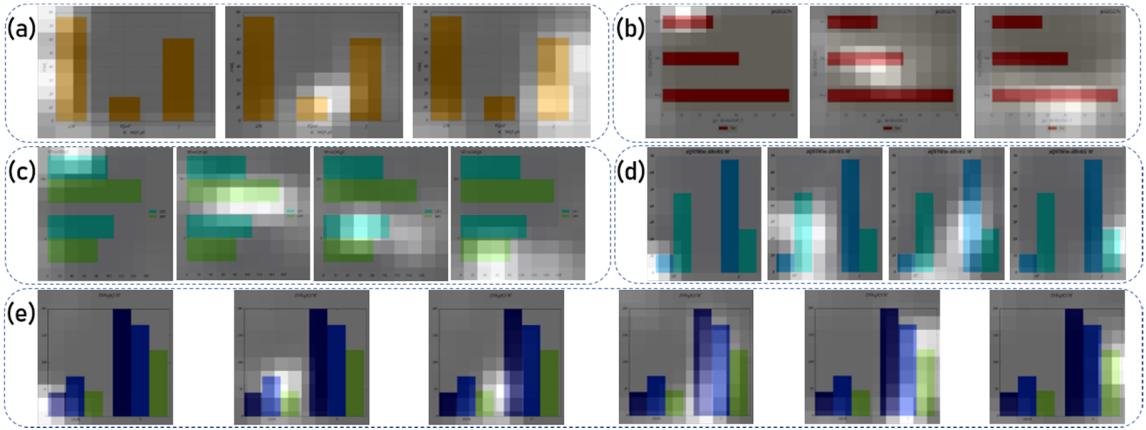

Figure 7. Visualizations of attention vectors and their bar charts. (a) to (e) cover horizontal and vertical bar charts and bar charts with 1 to 3 bars per series

To visually investigate how the attention mechanism works, we combine the attention vector with the original bar chart image. We first resize the attention vector back to two dimensions. Because the spatial correspondence is retained, the attention vector is then overlaid onto the original image. Figure 7 shows five examples from the synthetic test set. In these visualizations, lighter areas indicate more visual attention. During the extraction process, the attention model produces different attention vectors for each iteration. The attention vector places considerable visual attention around the bar for which the decoder generates the bar vector. For horizontal bar charts (Figure 7 (a), (d), and (e)), the visual attention moves from left to right; for vertical bar charts (Figure 7 (b) and (c)), the move follows a top down direction. The attention model also works properly for bar charts with different bars per series (Figure 7 (a), (d), and (e)). These visualizations suggest that the attention model works as we expected.

## 6 DISCUSSION

Although our method is demonstrated to be effective, some limitations still exist. The training is not end-to-end, and some works are repeated. In its current state, our method has two separate parts that are trained separately. Although both parts contain CNNs that work as feature extractors, we cannot reuse the features due to the separation. This condition leads to repeated work which increases the running time. A future work item is to use a single feature extractor for the two parts through engineering effort. Then, the two parts can also be trained end-to-end by combining their loss functions.

We observe several situations in which our method does not work very well: (1) the colors or heights of the neighboring bars are very close; (2) excessive bars are present in the charts; (3) some bars are very tall or short compared with the other bars. After investigating the synthetic dataset, we find that the above-mentioned situations are less common. We believe that the poor performance is caused by the deficiency of these rare training charts that are generated randomly. A future direction will be to control the randomness of synthetic data. Moreover, an extensive study of the real-world charts is required to generate a good synthetic dataset.

Our method only works for a subset of the real-world charts. Our synthetic data approximates only a subset of the real-world charts, and we have simplified assumptions regarding the bar charts to scope the research space. However, real-world charts are diverse and not subject to the assumptions. Our method is likely to work poorly on unseen charts. Because neural networks can learn from training data, we believe that our method can be extended to handle most real-world charts if sufficient labeled real-world charts are accessible.

Data recovery depends on the quality of OCR, such as the recognition of text and calculation of the mapping. When evaluating our method, we often observe situations in which our method works properly but the result is poor due to OCR recognition errors. These publicly available OCR services are designed for various scenarios. Thus, a better way is to train a character-level recognition model specifically for charts. This approach will be a future work item to further improve the performance of our method. For the benchmark dataset, most of the existing researches[7][8] are based on the experiments conducted by the charts synthesized by themselves and the charts downloaded from the Internet, so building a suitable benchmark data set is worth doing in the future, which will promote the development of related research.

Our method is designed to be general and flexible to ensure its easy extendibility to other types of charts. Take pie chart for example. We first prepare the training data, which should include the pie charts, text bounding boxes and classes, and pie sector descriptions. The textual information extractor is versatile, and no modification is required. For the numeric information extractor, we only modify the last output layer of the decoder to adapt to pie sector descriptions. Then, the two parts are trained using the pie chart dataset. Because pie charts scarcely contain chart coordinates, data recovery for numeric information can be omitted. In addition, the generalization of the matching between textual and numeric information in our method can be further improved. In our work, the textual information and the numeric information are extracted in the same order, such as from top to bottom and from left to right. But this extraction method in the same order will cause two problems. One is that if our model extracts more (less) textual or numeric information, the subsequent textual and numeric matching will be mismatched from the wrong place. The second is that when our model is extended to irregular charts for information extraction, the matching would be complicated. Because the legends and sectors of irregular charts (such as pie charts) could be arranged out of order, our extraction method will make the textual and numeric information mismatched from the beginning. Therefore, we need a more complicated method to deal with this issue. One of our future tasks is to design an additional model for matching the textual and the numeric objects. A feasible idea is to use feature maps, numeric object description vectors, and text object description vectors as the inputs of the matching model, and the outputs are the probabilities that they match each other.

## 7 CONCLUSION

This study proposes a neural network-based method to reverse-engineer bar charts. For textual information extraction, we improve the efficiency by using object detection model to localize and classify textual elements simultaneously. For numeric information extraction, we use an encoder–decoder framework with attention mechanism to achieve high accuracy and robustness. Synthetic and real-world datasets are used to train and evaluate our method. The evaluations demonstrate that our method is effective. Our method can be extended to reverse-engineer other types of charts through appropriate modifications.

## ACKNOWLEDGMENTS

This work was supported in part by the National Natural Science and Technology Fundamental Resources Investigation Program of China (No. 2018FY10090002), the National Natural Science Foundation of China (No. 61672538 and 61872388), and the Natural Science Foundation of Hunan Province (No. 2020JJ4758). The data sets and source codes of this work are available at Github: https://github.com/csuvis/BarchartReverseEngineering.